\documentclass{article}
\usepackage{spconf,amsmath,amssymb,graphicx}
\usepackage{float}
\usepackage{placeins}
\usepackage{dblfloatfix}
\usepackage[draft,bookmarks=false]{hyperref}

\title{Fix, Don't Retake: Physics-Simulated Training for Correcting Patient Positioning Errors in Dental Panoramic Radiographs}

\name{Sema Helali$^{1,3}$ \qquad Lina Abu Nada$^{2}$ \qquad Faleh Tamimi$^{3}$ \qquad Rafat Damseh$^{1,3}$}

\address{$^{1}$Department of Computer Science and Software Engineering,\\ College of Computing and Artificial Intelligence, United Arab Emirates University, UAE\\
$^{2}$Department of Oral and Craniofacial Health Sciences, College of Dental Medicine,\\ University of Sharjah, UAE\\
$^{3}$Faculty of Dental Medicine and Oral Health Sciences, McGill University, Canada}

\begin{document}
\maketitle

\begin{abstract}
Patient-positioning errors can distort panoramic
radiographs and lead to repeat examinations. This
study investigated artificial intelligence for correcting
these errors in existing images. A physics-based
simulator generated paired correctly positioned and
mispositioned panoramics from cone-beam computed
tomography volumes, with controlled individual and
combined positioning errors. Five approaches were
compared: U-Net, pix2pix, conditional diffusion,
an error-conditioned model, and a mixture of
error-specific specialists. Evaluation used image
similarity and geometric metrics across three test
sets. All approaches improved mean image similarity
and reduced geometric errors, with U-Net and the
two error-aware approaches outperforming pix2pix
and conditional diffusion.
\end{abstract}


\section{Introduction}

Panoramic radiography provides an overview of the teeth, jaws,
and surrounding structures and is widely used in dental
assessment and treatment planning. However, incorrect patient
positioning can affect the appearance of anatomical structures
and limit diagnostic interpretation. Historical studies classified
33\% and 24.9\% of the panoramic radiographs examined as
diagnostically unacceptable. Images that require repeated
examination expose patients to additional, potentially avoidable
radiation \cite{Rushton1999,Dhillon2012}.

Common positioning errors include placing
the patient too far forward or backward, raising or lowering
the chin, rotating or tilting the head, and failing to maintain
the tongue against the palate. These errors can cause blurring,
alter the apparent width of the anterior teeth and the appearance
of the occlusal plane, and produce asymmetry. Incorrect tongue
positioning creates a radiolucent band that can obscure the
maxillary tooth apices \cite{Dhillon2012,Panyarak2016}.

Artificial intelligence (AI) has been applied to the detection of
positioning errors and assessment of panoramic images
\cite{Hu2026}. Other studies have investigated generative
adversarial networks for restoring images affected by artificial
blur and noise \cite{Kim2023,Kolukisa2025}. These tasks differ
from correcting changes in projected anatomy caused by patient
positioning. More closely related work by Du et al. used a
convolutional neural network to estimate dental-arch positioning
errors and reconstruct panoramic images using corrected dental
curvature \cite{Du2018}. Further investigation is needed to assess
how different AI approaches perform when correcting individual
and combined positioning errors in panoramic images.

This study investigated whether AI could correct patient-positioning errors in existing panoramic radiographs, potentially reducing repeat examinations. To address the lack of suitable datasets, a physics-based simulator was developed to generate paired correctly positioned and mispositioned panoramics from CBCT volumes. These images were used to train and evaluate AI approaches for correcting individual and combined positioning errors.

\section{Methods}

\subsection{Paired Data Generation}

\textbf{CBCT preparation and reference positioning.}
CBCT volumes were obtained from the MMDental dataset \cite{MMDental2025}, with voxel intensities treated as approximate Hounsfield units (HU). The jaws and teeth were segmented using DentalSegmentator \cite{DentalSegmentator2024}. The segmentations were used to estimate the occlusal and mid-sagittal planes for consistent head alignment. The head was then positioned using the anterior incisal midpoint as a reference, with the occlusal plane inclined $7^\circ$ downward anteriorly. To establish the reference tongue position, oral-cavity air (HU $< -500$) was replaced with soft tissue (40~HU).

\textbf{Physics-based panoramic simulation.}
Panoramic radiographs were generated using a virtual X-ray source and narrow-slot detector rotating around the head along a curved focal trough. CBCT values were clipped to $[-1000,4000]$ HU and converted into linear attenuation coefficients:

\[
\mu(\mathbf{x})=\mu_{\mathrm{water}}\left(1+\frac{\mathrm{HU}(\mathbf{x})}{1000}\right),
\]

where $\mu_{\mathrm{water}}=0.020~\mathrm{mm}^{-1}$. X-ray transmission was calculated using the monochromatic Beer--Lambert approximation:

\[
I_k(\mathbf{d})=\exp\left(-\int_{\mathbf{s}_k}^{\mathbf{d}_k}\mu(\mathbf{x})\,d\ell\right),
\]

where $\mathbf{s}_k$ and $\mathbf{d}_k$ are the source and detector-pixel positions at acquisition step $k$. Ray integrals were evaluated at 0.5 mm intervals within the CBCT field of view.

At each scanning step, ray casting generated a vertical projection strip. The machine moved so that structures within the focal trough shifted by one detector pixel between steps. Time-delay integration (TDI) aligned the strips using integer column shifts and averaged their transmitted intensities. The final panoramic image was calculated as

\[
P(\mathbf{d})=-\ln\left(\frac{1}{N(\mathbf{d})}\sum_{k\in\mathcal{K}(\mathbf{d})}I_k(\mathbf{d}+\boldsymbol{\Delta}_k)\right),
\]

where $\mathcal{K}(\mathbf{d})$ contains the projections used for each pixel, $N(\mathbf{d})$ is their number, and $\boldsymbol{\Delta}_k$ is the column shift.

A validity mask excluded pixels outside the valid CBCT coverage from training and evaluation. Four virtual panoramic geometries (M1--M4) were defined using different imaging parameters. M1--M3 were used for training, validation, and testing, while M4 was reserved for evaluating generalization to a different machine geometry.

\textbf{Positioning-error generation.} 
Patient-positioning errors were simulated by applying rigid transformations to the CBCT anatomy while keeping the virtual X-ray machine fixed. Rotations were applied around the anterior bite point. The transformed volume was defined as

\[
V_{\theta}(\mathbf{x}) =
V_{\mathrm{ref}}(E_{\theta}^{-1}\mathbf{x}),
\qquad E_{\theta}=TR,
\]

where $V_{\mathrm{ref}}$ is the reference volume, $\mathbf{x}$ is a 3D coordinate, and $E_{\theta}$ combines translation $T$ and rotation $R$ according to the positioning-error parameters $\theta$.

\begin{figure*}[!t]
    \centering
    \includegraphics[width=\textwidth]{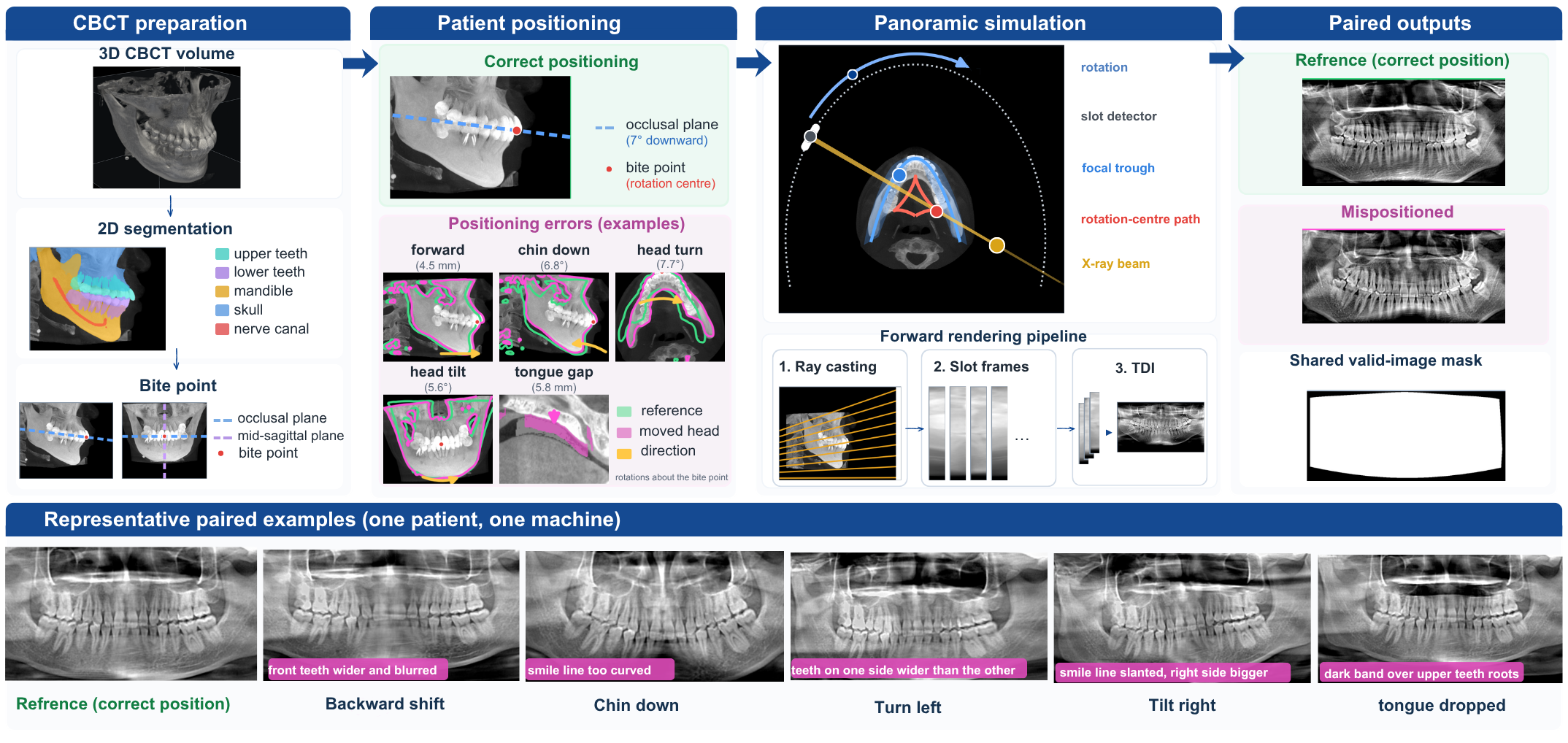}
    \caption{Overview of paired panoramic data generation
    from CBCT volumes, including positioning errors
    and physics-based panoramic simulation.}
    \label{fig:data_generation}
\end{figure*}

Five positioning-error types were simulated including forward/backward movement, chin up/down, head turn, head tilt, and tongue positioning \cite{Head, tongue}. Tongue-positioning errors were simulated by introducing an air gap ($-1000$ HU) beneath the hard palate. Errors were generated at three severity levels (Table~\ref{tab:errors}). Multiple positioning errors were also combined, excluding contradictory movements.
Figure~\ref{fig:data_generation} illustrates the
complete paired data generation process and
representative positioning errors.

\begin{table}[H]
\centering
\caption{Severity ranges of simulated positioning errors.}
\label{tab:errors}
\small
\setlength{\tabcolsep}{3pt}
\renewcommand{\arraystretch}{1.2}
\begin{tabular}{@{}lccc@{}}
\hline
\textbf{Error} &
\textbf{Mild} &
\textbf{Moderate} &
\textbf{Severe} \\
\hline
Forward/backward (mm) & 2--4 & 4--7 & 7--10 \\
Chin down/up ($^\circ$) & 3--6 & 6--10 & 10--15 \\
Head turn ($^\circ$) & 4.5--6 & 6--9 & 9--12 \\
Head tilt ($^\circ$) & 2--4 & 4--6 & 6--9 \\
Tongue air gap (mm) & 1.5--3 & 3--6 & 6--10 \\
\hline
\end{tabular}
\end{table}

\subsection{AI Correction Approaches}

Each of the five approaches received a mispositioned image $x$ and
validity mask $m$, with the correctly positioned image
$y$ used as the training reference.

\textbf{U-Net.}
A six-level residual U-Net \cite{unet} with
self-attention in its two deepest levels and
bottleneck was used for direct image correction.
The mispositioned image and validity mask were
provided as a two-channel input. Training minimized
the masked L1 loss:
\begin{equation}
\mathcal{L}_{1}^{m} =
\frac{\sum_i m_i |\hat{y}_i-y_i|}
{\sum_i m_i},
\label{eq:masked_l1}
\end{equation}
where $i$ indexes pixels, $\hat{y}$ is the corrected
image, and $m_i$ indicates valid pixels.

\textbf{pix2pix.}
The pix2pix approach \cite{pix2pix} used the same
U-Net architecture for its generator,
along with a PatchGAN discriminator. The generator was trained
using
\begin{equation}
\mathcal{L}_{G} =
\mathcal{L}_{\mathrm{adv}}
+100\,\mathcal{L}_{1}^{m},
\label{eq:pix2pix_loss}
\end{equation}
where $\mathcal{L}_{\mathrm{adv}}$ is the adversarial
loss calculated over valid image regions.

\textbf{Conditional diffusion.}
A conditional diffusion model \cite{HO} was
implemented using a time-conditioned U-Net.
The network received a noisy reference image,
the mispositioned image, and validity mask.
It was trained to predict diffusion velocity
\cite{Salimans2022} using a masked squared-error
loss and a cosine noise schedule \cite{Nichol2021}.
Corrected images were generated using DDIM sampling
\cite{Song2021}, starting from random noise and using
the predicted velocity to gradually reconstruct the
image.

\textbf{Error-conditioned U-Net.}
An error-prediction head was added to the U-Net
bottleneck to estimate the presence and magnitude
of nine positioning-error types (both directions of
the four geometric errors, plus tongue position). The resulting
18-dimensional vector conditioned the decoder
through FiLM layers \cite{Perez2018}. The model
was optimized using
\begin{equation}
\mathcal{L}_{\mathrm{int}} =
\mathcal{L}_{1}^{m}
+0.1\,(1-\mathrm{SSIM}^{m})
+0.1\,\mathcal{L}_{\mathrm{err}},
\label{eq:integrated_loss}
\end{equation}
where $\mathrm{SSIM}^{m}$ is the structural
similarity index \cite{Wang2004} calculated over
valid pixels, and $\mathcal{L}_{\mathrm{err}}$
combines binary cross-entropy for error presence
and smooth L1 loss for the magnitudes of present
errors. Ground-truth error information was
gradually replaced by predicted errors during
training. Only predicted errors were used during
testing.

\textbf{Mixture of error-specific specialists.}
Five U-Net specialists, one per positioning-error family, were fine-tuned from
the trained U-Net, following the mixture-of-experts concept \cite{Jacob}.
 Each was trained on images
containing its corresponding error, alone or
combined with others, using
$\mathcal{L}_{1}^{m}+0.1\,(1-\mathrm{SSIM}^{m})$.

Specialists were selected using error probabilities
predicted by the error-conditioned U-Net, with a
threshold of 0.5. Their outputs were averaged to produce
the final correction, with the trained U-Net
used when none was selected.

\section{Experiments and Results}
\subsection{Dataset and Splits}

The dataset comprised 12,340 image pairs from
358 patients, divided into training (283 patients,
8,490 pairs), validation (25 patients, 500 pairs),
and testing (50 patients, 3350 pairs), without patient overlap.

Three test sets used the same 50 test patients.
The natural-error set (1,000 pairs, M1--M3)
assessed correction under a sampled mixture of
individual and combined errors. The balanced-error
set (1,350 pairs, M1--M3) included all nine error
categories at three severity levels per patient,
with one error per image, enabling comparisons
across error types and severity. The M4 set (1,000 pairs) used the same error sampling as the natural-error set. It tested correction on a simulated imaging geometry that was not used
for training or validation.

\begin{table*}[!t]
\centering
\caption{Mean performance across the three test sets.
PSNR is in dB; incisor-width (IWE) and left/right (LRE)
errors are percentages. Higher ($\uparrow$) or lower
($\downarrow$) values are better. Best results are bold.}
\label{tab:overall_results}
\small
\setlength{\tabcolsep}{3pt}
\renewcommand{\arraystretch}{1.15}
\begin{tabular}{l*{12}{c}}
\hline
& \multicolumn{4}{c}{Natural}
& \multicolumn{4}{c}{Balanced}
& \multicolumn{4}{c}{Unseen geometry M4} \\
\cline{2-5}\cline{6-9}\cline{10-13}
Approach
& PSNR$\uparrow$ & SSIM$\uparrow$ & IWE$\downarrow$ & LRE$\downarrow$
& PSNR$\uparrow$ & SSIM$\uparrow$ & IWE$\downarrow$ & LRE$\downarrow$
& PSNR$\uparrow$ & SSIM$\uparrow$ & IWE$\downarrow$ & LRE$\downarrow$ \\
\hline
Uncorrected
& 18.81 & .577 & 5.57 & 5.99
& 16.64 & .458 & 5.91 & 6.62
& 17.90 & .550 & 5.32 & 6.73 \\
U-Net
& 23.25 & .703 & 2.64 & 2.04
& 21.59 & .659 & \textbf{1.78} & 1.93
& 21.25 & .649 & 3.10 & 2.93 \\
pix2pix
& 21.51 & .634 & 3.79 & 3.17
& 19.72 & .566 & 2.43 & 3.00
& 19.85 & .586 & 4.01 & 4.33 \\
Cond.\ diffusion 
& 21.18 & .624 & 3.62 & 3.30
& 19.32 & .565 & 2.68 & 2.76
& 19.31 & .562 & 3.96 & 3.62 \\
Error-conditioned
& \textbf{23.62} & .705 & 2.76 & 1.95
& 21.97 & .667 & 2.33 & 1.80
& 21.54 & .656 & 3.11 & 2.65 \\
Mixture of specialists
& 23.61 & \textbf{.709} & \textbf{2.28} & \textbf{1.64}
& \textbf{22.06} & \textbf{.673} & 2.07 & \textbf{1.67}
& \textbf{21.68} & \textbf{.660} & \textbf{2.89} & \textbf{2.44} \\
\hline
\end{tabular}
\end{table*}

\subsection{Experimental Setup}

Inputs were $512\times1024$; losses and metrics were computed
inside the valid mask. Training used a random intensity processing
per pair. U-Net, pix2pix, and the error-conditioned model were trained
for 25k steps and diffusion for ${\sim}$60k steps (AdamW, lr
$10^{-4}$, batch 6). Each specialist was fine-tuned from the U-Net
for 2.5k steps on training images containing its error (lr
$2\times10^{-5}$, batch 4). Checkpoints were selected on the validation set, while test sets were
used once. Diffusion used 25-step DDIM with one sample per input.

\begin{figure}[t]
\centering
\includegraphics[width=\columnwidth]{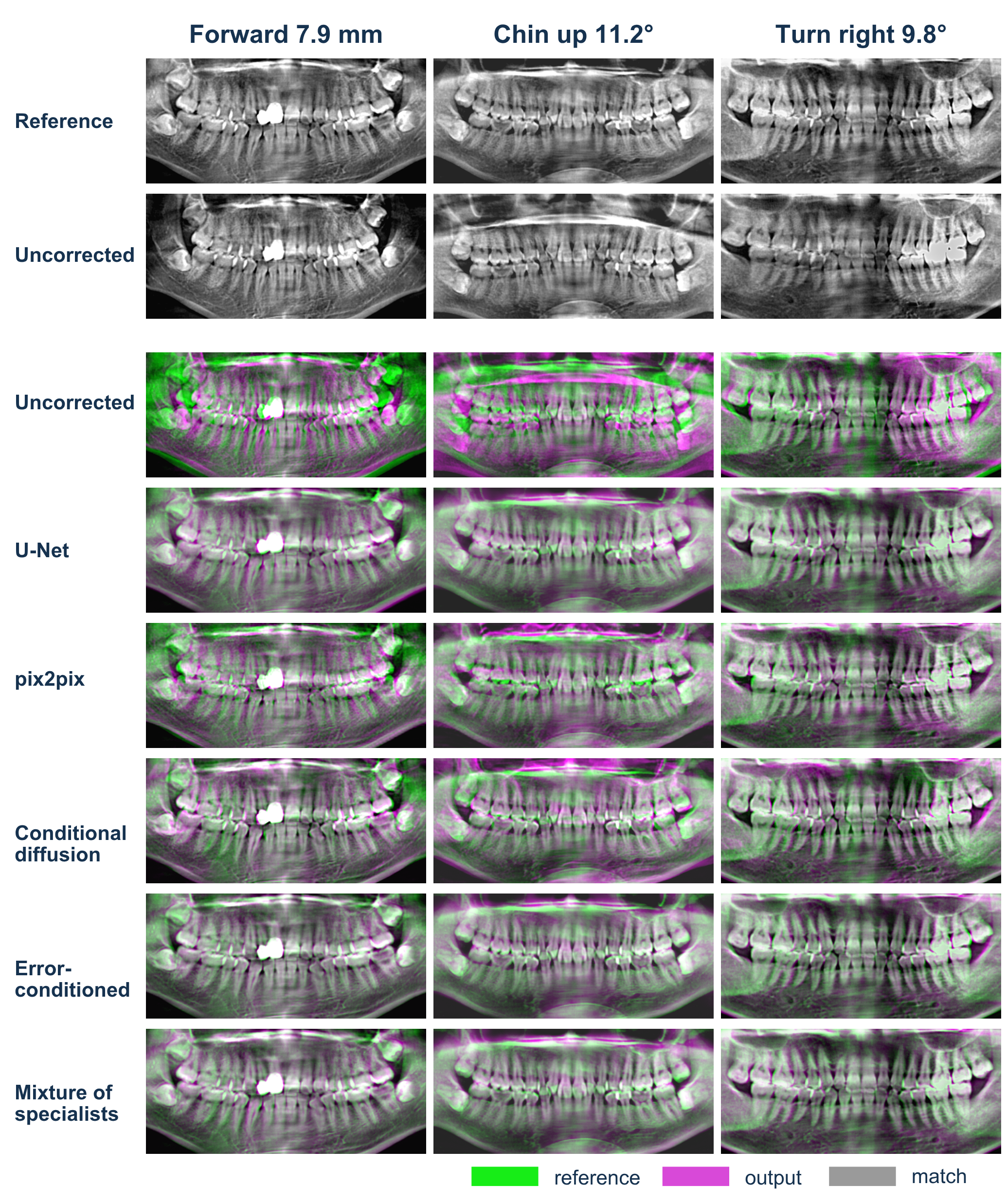}
\caption{Qualitative comparison for three positioning
errors. Overlays show the reference in green and the
compared image in magenta; gray indicates agreement.}
\label{fig:qualitative_results}
\end{figure}

\subsection{Evaluation Metrics}

\textbf{Image similarity.}
Peak signal-to-noise ratio (PSNR) measured differences
in pixel values, while the structural similarity index
measure (SSIM) measured similarity in local brightness,
contrast, and structure.

\textbf{Geometric errors.}
Two metrics assessed changes in front-teeth width
and differences in size between the left and right
sides. Each region was compared with the reference
at different widths, using normalized cross-correlation
to identify the best match.

Incisor-width error (IWE) was $100\,|s-1|\,\%$, where $s$
is the estimated front-teeth width in the corrected
image divided by that in the reference.
Left/right error (LRE) was $100\,|s_L/s_R-1|\,\%$, where
$s_L$ and $s_R$ are the corresponding width ratios
for the left and right posterior regions.

\subsection{Results}

All five approaches improved mean image similarity and reduced geometric errors across
the three test sets (Table~\ref{tab:overall_results}). U-Net and the two error-aware
approaches outperformed pix2pix and conditional diffusion across all four metrics.
Improvements were retained on M4, a simulated geometry excluded from training, where
the mixture achieved the best results.

All approaches improved PSNR at each severity level, with larger gains for more severe
errors (Table~\ref{tab:severity}).
\begin{table}[!t]
\centering
\caption{Mean results by severity on the balanced set
(450 images per level). Best results within each
level are bold.}
\label{tab:severity}
\small
\setlength{\tabcolsep}{2pt}
\renewcommand{\arraystretch}{1.1}
\begin{tabular}{llcccc}
\hline
Level & Approach & PSNR$\uparrow$ & SSIM$\uparrow$
& IWE$\downarrow$ & LRE$\downarrow$ \\
\hline
Mild & Uncorrected & 18.55 & .516 & 2.45 & 3.35 \\
& U-Net & 22.51 & .692 & 2.08 & 1.61 \\
& pix2pix & 20.77 & .616 & \textbf{2.01} & 2.28 \\
& Cond.\ diffusion & 20.35 & .607 & 2.54 & 1.68 \\
& Error-conditioned & 23.04 & .698 & 2.53 & \textbf{1.35} \\
& Mixture of specialists
& \textbf{23.15} & \textbf{.712} & 2.21 & 1.44 \\
\hline
Moderate & Uncorrected & 16.39 & .449 & 4.91 & 4.94 \\
& U-Net & 21.67 & .663 & \textbf{1.52} & 1.75 \\
& pix2pix & 19.75 & .566 & 1.86 & 2.84 \\
& Cond.\ diffusion & 19.32 & .566 & 2.65 & 2.50 \\
& Error-conditioned & 21.92 & .668 & 2.36 & 1.70 \\
& Mixture of specialists
& \textbf{22.05} & \textbf{.673} & 1.94 & \textbf{1.68} \\
\hline
Severe & Uncorrected & 14.99 & .408 & 10.37 & 11.56 \\
& U-Net & 20.57 & .623 & \textbf{1.76} & 2.42 \\
& pix2pix & 18.65 & .518 & 3.40 & 3.86 \\
& Cond.\ diffusion & 18.28 & .522 & 2.84 & 4.10 \\
& Error-conditioned & 20.96 & \textbf{.636} & 2.11 & 2.36 \\
& Mixture of specialists
& \textbf{20.97} & .632 & 2.06 & \textbf{1.89} \\
\hline
\end{tabular}
\end{table}

Figure~\ref{fig:qualitative_results} shows the same pattern. After correction, the
overlays were mostly grey, indicating closer alignment with the reference, with less
remaining misalignment for U-Net and the two error-aware approaches.

\section{Conclusion and Discussion}
This study developed a physics-based simulator to
generate paired panoramic images and compare AI
approaches for correcting positioning errors.
Experiments showed that these approaches could
reduce positioning-related distortions under
controlled simulated conditions. The findings
support investigating correction of existing images
as a potential alternative to repeating examinations.
Future work should evaluate real panoramic images
and include clinical assessment to determine whether
correction preserves the anatomy needed for diagnosis.
\section{Compliance with Ethical Standards}
This study used only the publicly available, de-identified MMDental dataset
\cite{MMDental2025}. Ethical approval was not required.

\section{Acknowledgments}
This work was supported by the United Arab Emirates University Strategic Research Program under Grant 12R315. AI tools assisted with coding and improving the manuscript's writing.

\FloatBarrier
\bibliographystyle{IEEEbib}
\bibliography{references}

\end{document}